\documentclass[10pt,twocolumn,letterpaper]{article}

\usepackage[pagenumbers]{cvpr} 

\usepackage{graphicx}
\usepackage{amsmath}
\usepackage{amssymb}
\usepackage{booktabs}
\usepackage{bm}
\usepackage{algorithm,algorithmic}
\usepackage{threeparttable}
\usepackage{booktabs, makecell}
\usepackage{array}
\usepackage{enumerate}
\usepackage{multirow}
\usepackage[pagebackref,breaklinks,colorlinks]{hyperref}

\usepackage[capitalize]{cleveref}
\crefname{section}{Sec.}{Secs.}
\Crefname{section}{Section}{Sections}
\Crefname{table}{Table}{Tables}
\crefname{table}{Tab.}{Tabs.}

\def\cvprPaperID{5434} 
\def\confName{CVPR}
\def\confYear{2022}

\begin{document}

\title{Synthetic Unknown Class Learning for Learning Unknowns}

\author{Jaeyeon Jang\\
Department of Industrial Engineering and Management Sciences, Northwestern University\\
Evanston, IL 60208 USA\\
{\tt\small jaeyeon.jang@northwestern.edu}
}
\maketitle

\begin{abstract}
   This paper addresses the open set recognition (OSR) problem, where the goal is to correctly classify samples of known classes while detecting unknown samples to reject. In the OSR problem, “unknown” is assumed to have infinite possibilities because we have no knowledge about unknowns until they emerge. Intuitively, the more an OSR system explores the possibilities of unknowns, the more likely it is to detect unknowns. Thus, this paper proposes a novel synthetic unknown class learning method that generates unknown-like samples while maintaining diversity between the generated samples and learns these samples. In addition to this unknown sample generation process, knowledge distillation is introduced to provide room for learning synthetic unknowns. By learning the unknown-like samples and known samples in an alternating manner, the proposed method can not only experience diverse synthetic unknowns but also reduce overgeneralization with respect to known classes. Experiments on several benchmark datasets show that the proposed method significantly outperforms other state-of-the-art approaches. It is also shown that realistic unknown digits/letters can be generated and learned via the proposed method after training on the MNIST dataset.
\end{abstract}

\section{Introduction}
\label{sec:intro}
In the past few years, the performance of recognition systems has greatly improved thanks to advancements in deep learning~\cite{He2016, Simonyan2015, Tan2019}. However, the vast majority of recognition systems still have difficulty detecting unknown samples because their performance improvements are obtained with the closed-world assumption, in which all categories are known a priori. As more applications are encountered under changing and open environments~\cite{Wong2019, Rocha2017, Jang2020a, Henrydoss2017}, the robustness of pattern recognition models to open environments where knowns and unknowns coexist becomes more important. In an open environment, we should reject unknown samples that are unseen in the training stage while maintaining high classification performance on known class samples. This more complex but realistic recognition task is called open set recognition (OSR)~\cite{Scheirer2013}.

\begin{figure}[b]
  \centering
   \includegraphics[width=\linewidth]{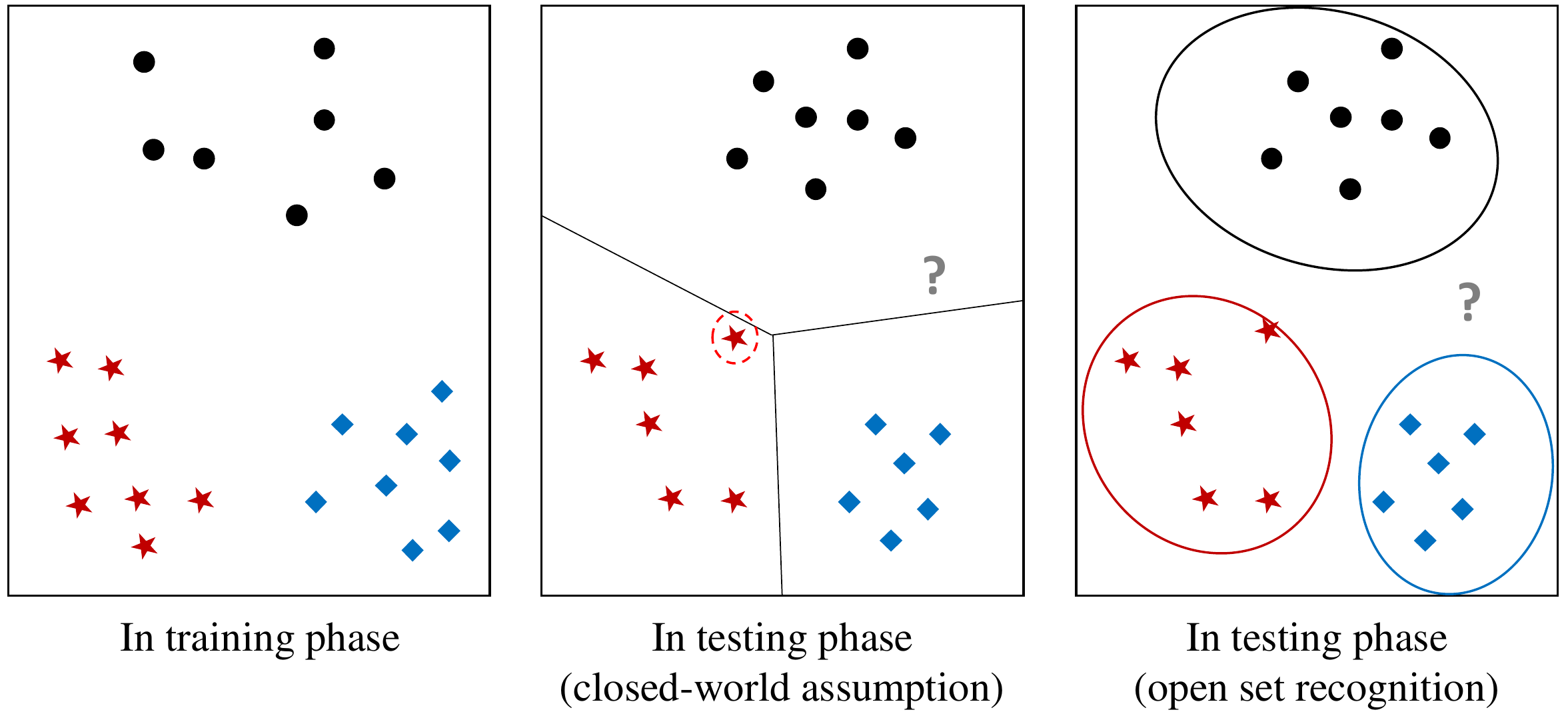}

   \caption{Difference between closed-set classification and OSR. With the closed-world assumption, overly generalized decision boundaries determine the unknown sample “?” as belonging to the black circle class, even though the decision boundaries can achieve high classification performance on known samples. However, with the concept of OSR, the unknown sample “?” is distinguished from all of the known classes.}
   \label{Fig_1}
\end{figure}

One of the most straightforward solutions for the OSR problem is to set a threshold on the posterior probability of the most likely class produced by classifiers; this is often called the confidence score~\cite{Hendrycks2017}. We can easily determine a sample as unknown if its confidence score is lower than the threshold. However, most classifiers, particularly deep neural networks (DNNs) with softmax output activation, also give very high confidence scores to unknown samples. This is because most neural networks used for classification are trained with the objective of maximizing the probability of belonging to one of the known classes based on the closed-world assumption, thereby establishing generalized decision boundaries~\cite{Jang2020b}. If the closed-world assumption holds, setting generalized decision boundaries can reduce the risk of misclassification~\cite{Moosavi-Dezfooli2019}. However, in an open environment, closed-world classifiers easily miss unknown samples because they classify the unknown samples into one of the known classes. For example, in~\cref{Fig_1}, the unknown sample “?” is misclassified as belonging to the black circle class even though the overly generalized decision boundary can correctly classify the red star into the red dashed circle that is far from the other red star samples. On the other hand, a well-trained OSR model can reject the unknown sample “?” while maintaining its generalization capability for known samples (high closed-set classification performance).

Researchers have found that information extracted from deep discriminative models, including general softmax convolutional neural networks (CNNs), is not sufficient for distinguishing unknowns. Thus, some studies have proposed methods to harness information from reconstructive networks for unknown detection to supplement the discriminative information for known classes learned by a softmax CNN~\cite{Yoshihashi2019, Spigler2019, Oza2019, Perera2020}. Here, unknown detection is a binary classification problem that involves determining whether a sample belongs to a known class or not. Despite the performance improvements achieved, these reconstructive networks are also vulnerable to overgeneralization since they do not learn discriminative representations for unknowns~\cite{Sun2020}, nor do they make a significant difference in reconstruction error~\cite{Jang2021}. That is, little discriminative information is given for the detection of unknowns. Other researchers have proposed utilizing generative networks to generate synthetic unknown samples and learn these unknown samples~\cite{Jang2021, Ge2017, Neal2018}. However, the synthetic samples generated by these works are not sufficiently diverse to cover unknowns because the existing methods do not consider diversity in the generated samples, or they limit the generated samples to a subspace of the known classes.

To overcome this shortcoming, this paper proposes a novel OSR method by assuming that unknowns come from various categories. The proposed method consists of training three different models. First, a teacher network, a general softmax CNN, is pretrained to classify only known class samples. Then, the knowledge learned by the teacher is transferred to a student network by using a knowledge distillation technique~\cite{Hinton2015}. Several nodes are added to the final layer of the student network (in addition to the nodes of known classes) to represent synthetic unknown classes. With this knowledge distillation technique, the student can not only learn softened probabilities about known classes but also leave room to learn unknowns. To help the student learn synthetic unknown classes, a recommender, \textit{i.e.}, a generative adversarial network (GAN) in which the generator takes the synthetic unknown class vector as a condition, generates samples for each synthetic unknown class and recommends these samples to the student. This process is defined as synthetic unknown class learning in this paper. The recommender can maintain the diversity of the generated samples since it tries to generate different samples for each synthetic unknown class. By alternately learning known samples through knowledge distillation and the synthetic unknown samples given by the recommender, the student can correctly detect unknown samples while minimizing the loss of classification performance on known samples.

Extensive experiments are conducted on various benchmark datasets, and the experimental results show that the proposed method outperforms other state-of-the-art methods in terms of unknown detection performance (the performance of classifying known/unknown) and OSR performance (the performance with respect to detecting unknowns and classifying known samples). Both knowledge distillation and synthetic unknown class learning contribute to the achieved performance improvement. Interestingly, it is also shown that the recommender can produce realistic fake samples that are almost visually unknown digits/letters.

This paper makes the following contributions:
\begin{itemize}
\item Knowledge distillation~\cite{Hinton2015}, a transfer learning technique, is employed to reduce overgeneralization on known classes by softening the posterior probabilities for known classes. This technique also helps the student and the recommender by leaving room for learning synthetic unknown classes.
\item Novel synthetic unknown class learning improves the OSR performance of the proposed network by generating realistic unknown-like samples and learning them.
\item Since the recommender is trained to produce different samples for each synthetic unknown class, the diversity in the synthetic unknown samples is maintained. In addition, an experiment shows that the generated synthetic samples have high intraclass diversity as well as high interclass diversity.
\end{itemize}

\section{Related Work}
\label{sec:related}
OSR can be seen as a problem of establishing decision boundaries to discriminate open set samples that are far from any known training samples while minimizing the loss of classification performance on known class samples. In summary, OSR consists of two main tasks: unknown detection and closed-set classification. Early works on OSR focused on setting decision boundaries to minimize the open set risk concept formalized in~\cite{Scheirer2013}, which combines the risk of misclassifying an unknown as a known sample with the risk of misclassifying a known sample as unknown, by using shallow machine learning models. For example, Scheirer \etal~\cite{Scheirer2013} used a set of one-vs.-set machines that expanded the one-vs.-rest support vector machine (SVM) by introducing an open set risk minimization term to SVM parameter optimization. Similarly, Scheirer \etal~\cite{Scheirer2014} adopted statistical extreme value theory (EVT) to establish more sophisticated decision boundaries based on the distribution of extreme scores produced by a radial basis function SVM.

With the recent advancements in deep learning, especially for classification tasks, many deep learning techniques have been introduced for OSR. The generalization capabilities of DNN classifiers provides high-performance classification enabling them to assign a test sample to the best-matching known class~\cite{Kawaguchi2017}. However, in an open environment in which unknowns can emerge, the generalization capabilities of these models can increase the risk of misclassifying an unknown sample as belonging to a known class. It is known that DNNs are vulnerable to overgeneralization, giving very high confidence scores to alien examples from well outside the distribution of the training examples~\cite{Kardan2017}. Thus, the main issue in OSR model development is how to mitigate overgeneralization while minimizing the loss of generalization capacity necessary for closed-set classification.

Many studies have utilized single DNNs to extract discriminative information for unknown detection. The first attempt in this category was the OpenMax model, which apportions known classes’ membership probabilities to an additional “unknown” class based on EVT~\cite{Bendale2016}. Shu \etal~\cite{Shu2017} found that replacing a softmax output layer with a sigmoid layer could help establish tighter decision boundaries. Similarly, Jang and Kim~\cite{Jang2020b} introduced a one-vs.-rest network architecture for the DNN output layer and combined class-specific decisions to obtain a robust unknown detection score. Yang \etal~\cite{Yang2020} proposed a convolutional prototype network to learn several prototypes for each known class while leaving room for unknowns in the feature space.

Reconstructive models have also been used for OSR for auxiliary purposes to calibrate classification decision scores or to implement unknown detection based on reconstruction error. The methods in this category also use softmax DNNs for closed-set classification. For example, Yoshihashi \etal~\cite{Yoshihashi2019} proposed a deep hierarchical reconstruction network that implements classification and reconstruction together and additionally considers feature representations in OpenMax score calculations. Oza and Patel~\cite{Oza2019} produced unknown detection scores by modeling the tail of the reconstruction error distribution based on EVT. Sun \etal~\cite{Sun2020} proposed a conditional Gaussian distribution learning (CGDL) method to learn a class-conditional distribution in the feature space by using a variational autoencoder. They used the distribution information in the feature space, as well as the reconstruction error of the variational autoencoder, for unknown detection.

Data augmentation strategies involving GANs have been used to generate synthetic unknown samples during training. For example, Ge \etal~\cite{Ge2017} used synthetic samples generated by a conditional GAN, which takes one-hot encoded class labels as conditions, to enhance the OpenMax model. However, a conditional GAN can only produce samples that can be inferred with class label combinations. Neal \etal~\cite{Neal2018} proposed a method to generate counterfactual unknown images that were close to known samples in the feature space to establish tight decision boundaries. However, this method ignores the possibility that an unknown sample far from the known samples in the feature space can fool the DNN classifier. Jang and Kim~\cite{Jang2021} trained a GAN that fooled a CNN classifier and then fed the generated samples to the CNN classifier to help it learn unknowns. However, the GAN produced synthetic examples that were similar to each other during each training epoch. Thus, this paper aims to provide a method that generates diverse unknown-like samples with the intuition that the DNN classifier can achieve high OSR performance if it can learn various probabilities for unknowns.

\section{Proposed Method}
\label{sec:method}
\begin{figure*}
  \centering
  \begin{subfigure}{0.4\linewidth}
    \includegraphics[width=\linewidth]{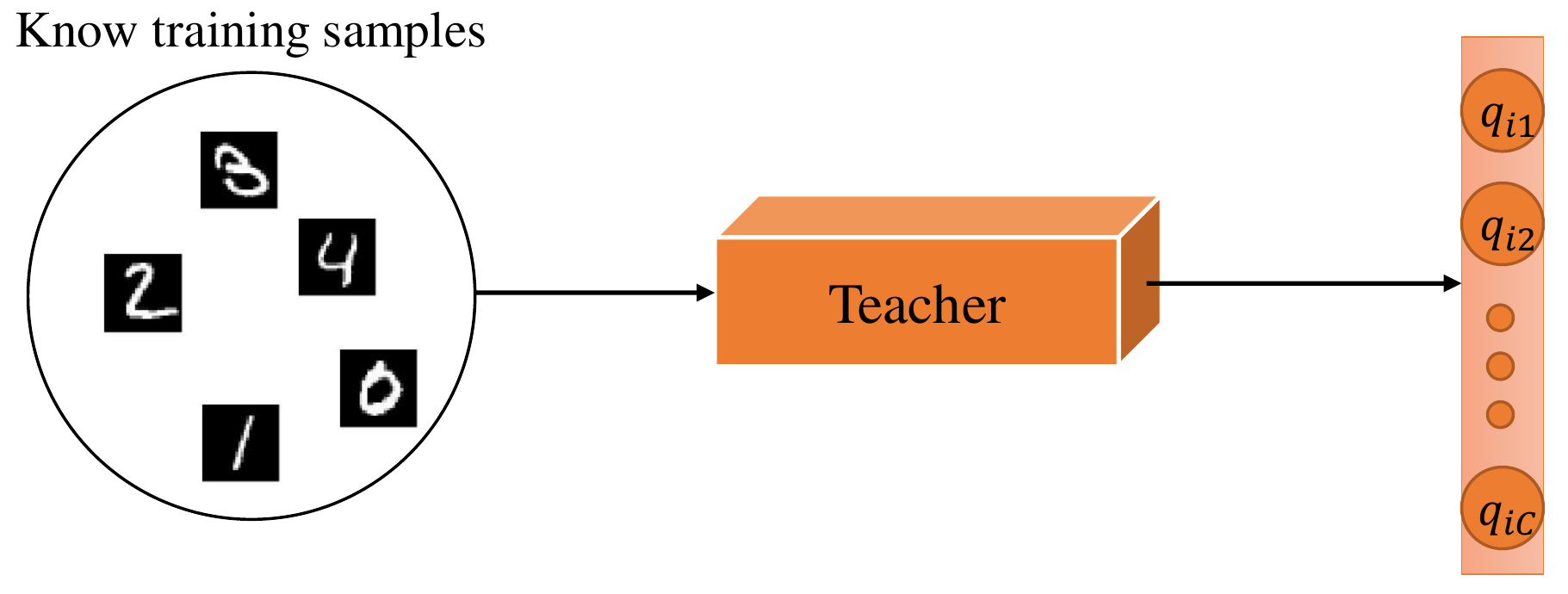}
    \caption{Step 1: Pretraining the teacher}
  \end{subfigure}
  \hfill
  \begin{subfigure}{0.52\linewidth}
    \includegraphics[width=\linewidth]{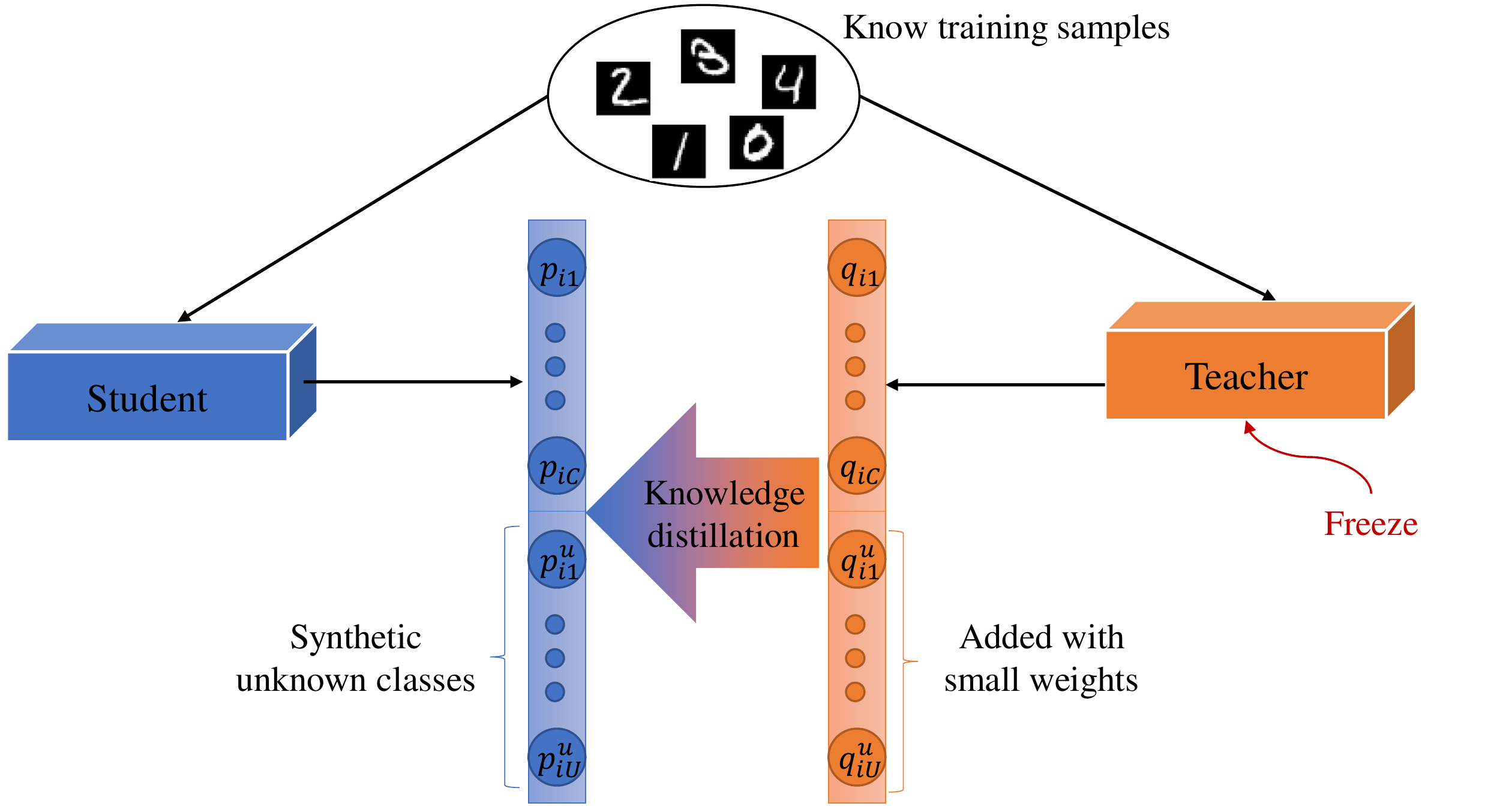}
    \caption{Step 2-1: Knowledge distillation}
  \end{subfigure}
    \begin{subfigure}{0.75\linewidth}
    \includegraphics[width=\linewidth]{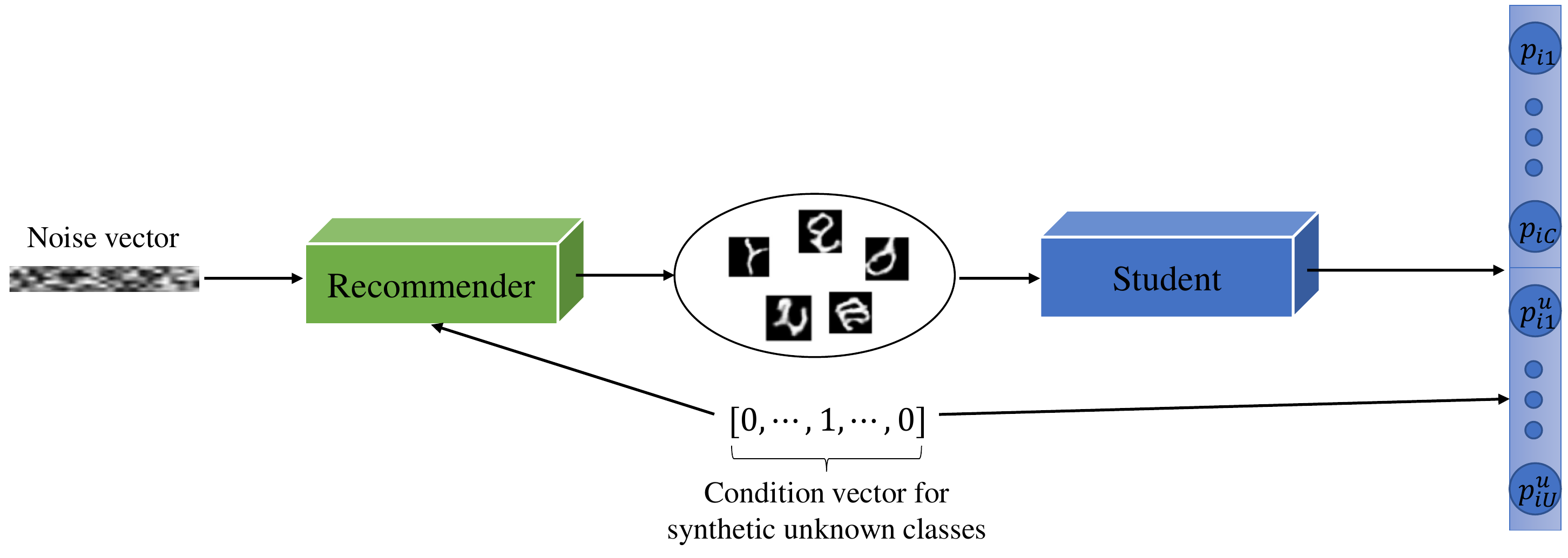}
    \caption{Step 2-2: Synthetic unknown class learning}
  \end{subfigure}
  \caption{Overview of the proposed method. $\bm{q}_{ik}$ and $\bm{p}_{ik}$ indicate the posterior probabilities of the sample-class label pair $(\bm{x}_i,y_i)$ for the $k$-th class of the teacher and the student, respectively. Similarly, $\bm{q}_{ik}^u$ and $\bm{p}_{ik}^u$ are the $k$-th synthetic unknown class membership probabilities. $C$ and $U$ denote the number of known classes and the number of synthetic unknown classes, respectively.}
  \label{Fig_2}
\end{figure*}

\cref{Fig_2} shows an overview of the proposed method. The proposed method is comprised of three different training steps. First, the teacher network, a general softmax CNN, is pretrained with known training samples. Then, two different complementary substeps are implemented alternately. The first substep transfers the teacher’s knowledge regarding known training samples by harnessing a knowledge distillation technique~\cite{Hinton2015}, which is often called teacher-student learning. Before this substep, the student network’s weights for synthetic unknown classes are initialized with small values, and the synthetic unknown classes are added to the teacher with small (negligible) weights. In the second substep, the student network is trained on the synthetic unknown samples provided by the recommender. Here, the recommender attempts to produce samples that can fool the discriminator into determining the samples as “real” and can enable the student to determine that the samples belong to a synthetic unknown class (given as condition vectors). Finally, during the testing phase, the teacher’s knowledge and the student’s knowledge are combined to make decisions.

\subsection{Knowledge distillation}
Following the conventional DNN classifier training scheme, the teacher network ($T$) is first trained with the aim of minimizing the cross-entropy loss function. DNN classifiers are usually overgeneralized, producing very high confidence scores even for unknowns during testing. Thus, we introduce an additional network, the student network, and apply a knowledge distillation technique~\cite{Hinton2015} based on the pretrained teacher network to relax the posterior probabilities of the student network for known classes. Knowledge distillation transfers softened versions of the target probabilities provided by the pretrained teacher network. The soft target probabilities are defined by introducing a temperature $\tau$ in the softmax function as follows:
\begin{equation}
\begin{aligned}
  &q_{ic}^\tau=\frac{\text{exp}(l_{ic}/\tau)}{\sum_{k=1}^C\text{exp}(l_{ik}/\tau)+\sum_{k=1}^U\text{exp}(l_{ik}^u/\tau)}\;\text{and}\\ &q_{ic}^{u\tau}=\frac{\text{exp}(l_{ic}^u/\tau)}{\sum_{k=1}^C\text{exp}(l_{ik}/\tau)+\sum_{k=1}^U\text{exp}(l_{ik}^u/\tau)},
  \label{Eq_1}
\end{aligned}
\end{equation}
where $l_{ik}$ is the logit of sample $\bm{x}_i$ for the $k$-th known class and $l_{ik}^u$ is the logit for the $k$-th synthetic unknown class. The same temperature scaling technique is applied to the student network when it learns known training samples, yielding temperature-scaled posterior probabilities $p_{ic}^\tau$ and $p_{ic}^{u\tau}$. The student network is then trained to minimize the following knowledge distillation loss function:
\begin{equation}
  \mathcal{L}_{KD}(\mathcal{D}) = \frac{1}{|\mathcal{D}|}\sum_{(\bm{x}_i,y_i)\in\mathcal{D}}\mathcal{H}(Q_i^\tau,P_i^\tau),
  \label{Eq_2}
\end{equation}
where $\mathcal{D}$ is a known training sample set, $|\mathcal{D}|$ is the size of $\mathcal{D}$, $\mathcal{H}$ is the cross-entropy function, and $Q_i^\tau$ and $P_i^\tau$ are temperature-scaled probability vectors for the teacher and student, respectively.

Unlike original knowledge distillation~\cite{Hinton2015}, we do not consider the cross-entropy between the true labels and the nonscaled probability vector of the student because our purpose is to minimize overgeneralization while leaving room for learning unknowns. In this way, the student can have a chance to produce high probabilities for the synthetic unknown classes when evaluating an unfamiliar sample by reducing the risk of overfitting for known classes. This chance is further amplified via the synthetic unknown class learning approach described in the next subsection.

\subsection{Synthetic unknown class learning}
\begin{figure}[h]
  \centering
   \includegraphics[width=\linewidth]{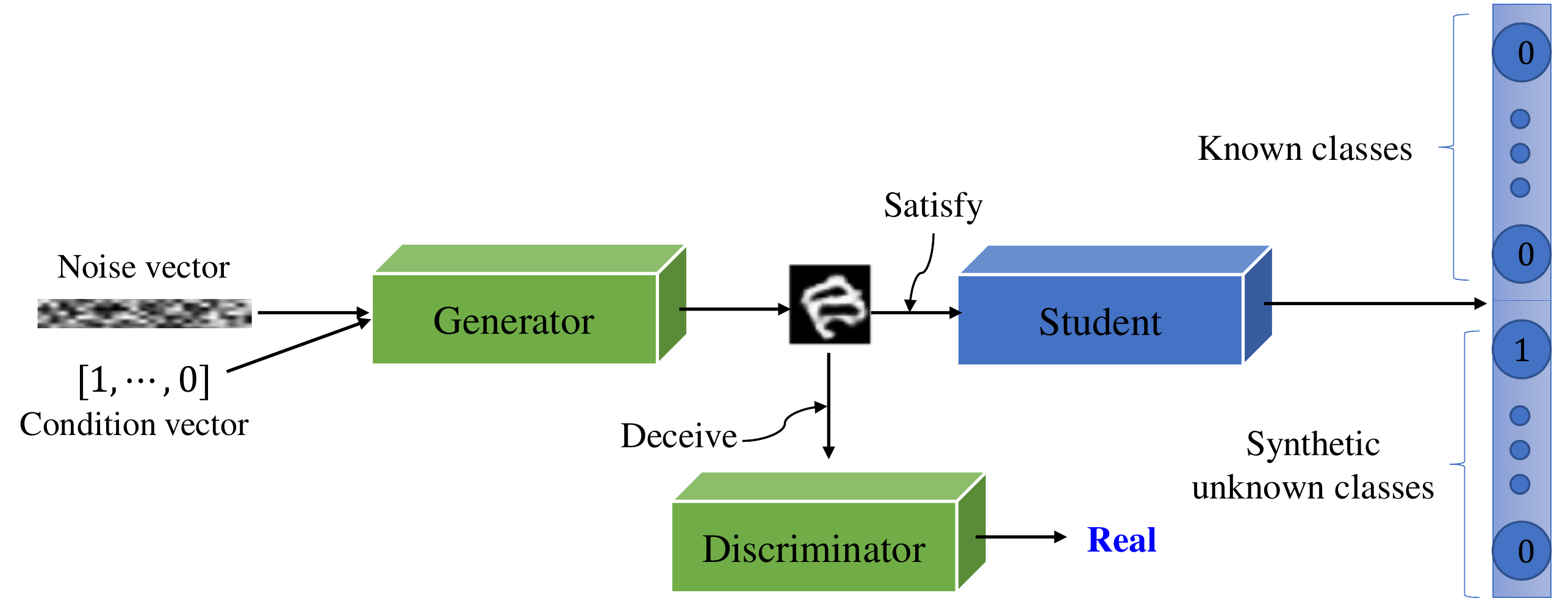}
   \caption{The training goal of the generator in the recommender is to produce samples that satisfy the student and deceive the discriminator.}
   \label{Fig_3}
\end{figure}

The recommender adopts a GAN structure consisting of a generator and a discriminator. The original GAN was trained with the aim of generating realistic samples by training the generator to deceive the discriminator, which was trained to discriminate between real and fake samples. In addition to this original goal, the generator in the recommender must satisfy the student by producing and recommending samples that follow the distribution of the target synthetic unknown class in the feature space of the student, as shown in~\cref{Fig_3}. Let $S$, $G$, and $D$ be the student, generator, and discriminator, respectively, and $\bm{\theta}_S$, $\bm{\theta}_G$, and $\bm{\theta}_D$ be the corresponding sets of parameters. Let us assume that a latent noise vector $\bm{z}$ follows a prior distribution $P_{pri}(\bm{z})$. Then, the generator is trained to minimize the following objective function:
\begin{equation}
    \min_{\bm{\theta}_G}\mathbb{E}_{\bm{z}\sim p_{\text{pri}}(\bm{z})}[\text{log}(1-D(G(\bm{z})))+\alpha\mathcal{H}([\bm{0},\bm{cv}], S(G(\bm{z})))],
  \label{Eq_3}
\end{equation}
where $[\bm{0},\bm{cv}]$ is the vector combining the zero vector for known classes and the randomly given condition vector $\bm{cv}$, and $\alpha$ is a balancing parameter. The discriminator is trained with the original goal of discriminating between real training samples and generated fake samples based on the objective function, which is defined as:
\begin{equation}
    \max_{\bm{\theta}_D}\mathbb{E}_{\bm{x}\sim \mathcal{D}}[\text{log}(D(\bm{x}))]+\mathbb{E}_{\bm{z}\sim p_{\text{pri}}(\bm{z})}[\text{log}(1-D(G(\bm{z})))].
  \label{Eq_4}
\end{equation}

Whenever the generator generates fake unknown samples, the generated samples are recommended to the student. Then, the student learns only the unknown samples that are worth learning based on the loss defined in \eqref{Eq_5}. The student determines that a sample is worth learning if the teacher assigns it a confidence score lower than $\lambda$. Here, $\lambda$ is the minimum confidence score threshold, which is lower than the confidence scores of the vast majority of the known training samples. The student network is trained on both the known training samples and recommended synthetic unknown samples in an alternating manner. The pseudocode in \cref{alg_1} shows the alternating training process.
\begin{equation}
    \mathcal{L}_S(\bm{z})=\mathcal{H}([\bm{0},\bm{cv}], S(G(\bm{z}))).
  \label{Eq_5}
\end{equation}
\begin{algorithm} 
 \caption{Pseudocode for alternating training}
\label{alg_1}
 \begin{algorithmic}[1]
 \renewcommand{\algorithmicrequire}{\textbf{Require:} }
 \REQUIRE Pretrained teacher network, student network $S$ with $\bm{\theta}_S$, generator $G$ with $\bm{\theta}_G$, discriminator $D$ with $\bm{\theta}_D$, known training dataset $\mathcal{D}$;
  \STATE Set batch size $N$;
 \WHILE{$\bm{\theta}_S$ has not converged}
  \STATE Sample $B_{\mathcal{D}}=\{\bm{x}_{(1)}, \bm{x}_{(2)}, \cdots, \bm{x}_{(N)}\}$ and produce corresponding temperature-scaled target probabilities;
  \STATE Sample $B_{\text{pri}}=\{\bm{z}_{(1)}, \bm{z}_{(2)}, \cdots, \bm{z}_{(N)}\}$ with the corresponding condition vector set;
  \STATE Update $\bm{\theta}_D$ based on \eqref{Eq_4};
  \STATE Update $\bm{\theta}_G$ based on \eqref{Eq_3};
  \STATE Update $\bm{\theta}_S$ by using $B_{\mathcal{D}}$ based on \eqref{Eq_2} and by using $B_{\text{pri}}$ based on \eqref{Eq_5};
\ENDWHILE
 \end{algorithmic} 
 \end{algorithm}

\subsection{Testing}
For testing, two strategies are proposed by using both the student’s knowledge, which includes both knowns and synthetic unknowns, and the teacher’s knowledge, which is specialized for knowns. The first strategy focuses on producing a score for unknown detection, leaving the closed-set classification task to the teacher network. Let $C$ and $U$ be the numbers of known classes and synthetic unknown classes, respectively. Then, the corresponding unknown score $\mathcal{U}(\bm{x}_i)$ is defined as follows:
\begin{equation}
    \mathcal{U}(\bm{x}_i)=(1-\max_{k\in\{1,\cdots,C\}}q_{ik})\times\sum_{k=1}^U p_{ik}^u.
  \label{Eq_6}
\end{equation}
The first strategy determines that sample $\bm{x}_i$ belongs to an unknown class if $\mathcal{U}(\bm{x}_i)>\delta$. Here, $\delta$ is the unknown score threshold for unknown detection.

As the second strategy, we propose setting a classwise decision boundary based on the average known probability score $\mathcal{K}(\bm{x}_i)_k=\frac{q_{ik}+p_{ik}}{2},\; \forall k \in\{1,\cdots,C\}$. With this known class score, the second strategy determines $y_i$, the class of $\bm{x}_i$, based on the following recognition rule:
\newcommand{\argmax}{\arg\!\max}
\begin{equation} 
y_i= \begin{cases}
\argmax_{k \in \{1,\cdots,C\}}\mathcal{K}(\bm{x}_i)_k & \text{if } \mathcal{K}(\bm{x}_i)_k > \epsilon_k
\\
    \text{"unknown"} & \text{otherwise}
  \end{cases}, \label{Eq_7}
\end{equation}
where $\epsilon_k$ is the known score threshold defined for a class $k$. In this paper, $\epsilon_k$ is set to ensure that 90\% of the class $k$ training samples are classified as belonging to class $k$.

\section{Experiments}
\label{sec:exp}
\subsection{Implementation details}
We employed a plain CNN as the backbone of the teacher network and the student network for training on the MNIST~\cite{Web2012} and the redesigned VGGNet for training on other image datasets (both networks are defined in~\cite{Yoshihashi2019}). The number of synthetic unknown classes was set to 10. The architecture of the recommender network can be found in our Supplementary Material. All networks were trained using the Adam optimizer with a learning rate of 0.002. The temperature $\tau$ in~\eqref{Eq_2} was set to 5, and the balancing parameter $\alpha$ in~\eqref{Eq_3} was set to 0.5. The sensitivity analysis results for $\tau$ and $\alpha$ can be found in our Supplementary Material. We set the minimum confidence score threshold $\lambda$, which ensured that 99\% of the training data had higher confidence scores than the threshold.

\subsection{Effects of recommended samples}
\begin{figure}[h]
  \centering
   \includegraphics[width=\linewidth]{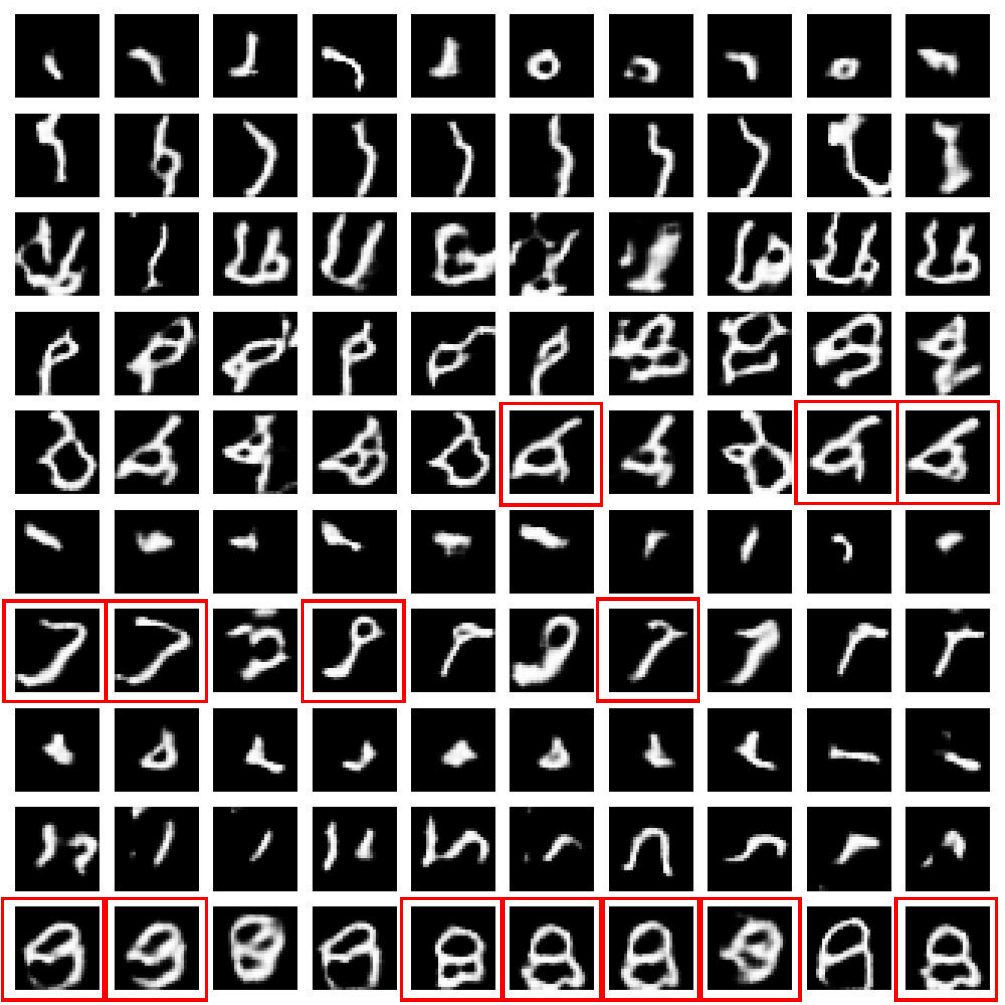}
   \caption{The synthetic samples generated by the recommender network. For each row, a different synthetic unknown class was activated in the condition vector.}
   \label{Fig_4}
\end{figure}

The student network learns unknowns through the synthetic samples generated by the recommender. Thus, it is important to validate whether the recommender produces good samples and whether the samples help the student learn unknowns. For this experiment, the MNIST dataset was split into six known classes (0–5) and four unknown classes (6–9), and only the training dataset containing the known classes was used for training. \cref{Fig_4} shows that the generator in the recommender produced a diverse set of unknown-like samples according to given synthetic unknown classes. In addition, we can see that the synthetic samples had high intraclass diversity and interclass diversity. Interestingly, some generated samples look like unknown numbers “6”, “7”, “8”, and “9” (see the samples in red boxes). In summary, this figure shows that our primary goal of obtaining diverse synthetic unknown samples could be achieved by the proposed learning method.

\begin{figure}[h]
  \centering
   \includegraphics[width=0.85\linewidth]{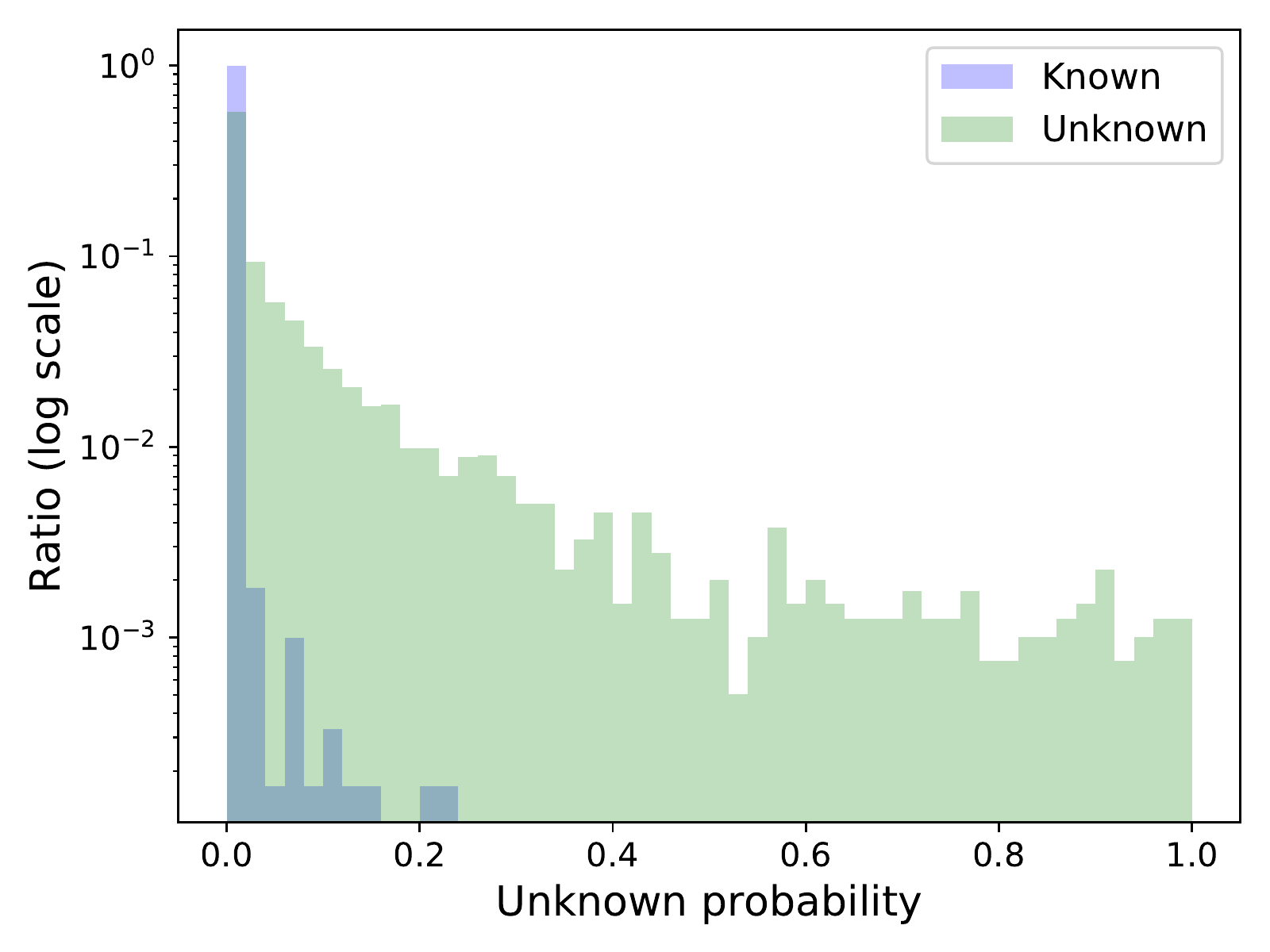}
   \caption{Unknown probability histogram for known and unknown samples. Here, the unknown probability is the sum of the posterior probabilities for the synthetic unknown classes.}
   \label{Fig_5}
\end{figure}

We produced an unknown probability, which is the sum of the posterior probabilities for synthetic unknown classes, for both the known and unknown test samples. \cref{Fig_5} shows that the student could provide much higher unknown probabilities to unknowns than to knowns after learning the unknown-like samples produced by the recommender. This reveals that the student’s knowledge about synthetic unknowns could be utilized for unknown detection.

\subsection{Ablation study}
In this section, the effect of each component in the proposed method was analyzed. All the baselines applied the second recognition strategy of setting class-specific decision boundaries. Specifically, the following four baselines were compared:
\begin{enumerate}[1)]
\item T: This baseline uses only the teacher network, a general softmax CNN.
\item TS: A pretrained teacher network transfers its knowledge through a knowledge distillation technique to train the student network. However, no synthetic unknown sample is fed during training.
\item RS: In this baseline, the recommender provides synthetic unknown samples to the student for unknown learning. However, the recommender and the student start the learning process from scratch without any prior knowledge provided by the teacher.
\item TRS: This baseline is the proposed method that uses the teacher, the recommender, and the student together.
\end{enumerate}

For an ablation study, the 10 digit classes of the MNIST dataset were used as known classes, and the 47 letter classes of the Extended MNIST (EMNIST) dataset~\cite{Cohen2017} were considered “unknown”. We set various open set scenarios by varying the number of unknown letter classes used in the scenarios, and we then measured how open each scenario was by using the openness concept, which is defined as follows:
\begin{equation}
    \text{openness}=1-\sqrt{\frac{2C_{TR}}{C_{TE}+C_{R}}},
  \label{Eq_8}
\end{equation}
where $C_{TR}$ is the number of classes used in training, $C_{TE}$ is the number of classes used in testing, and $C_{R}$ is the number of classes to be recognized during testing. The openness was varied from 4.7\% to 45.4\% by randomly sampling 2 to 47 unknown classes in intervals of 5. For each scenario, the macroaveraged F1 score over the known classes and “unknown” was used for evaluation.

\begin{figure}[h]
  \centering
   \includegraphics[width=0.85\linewidth]{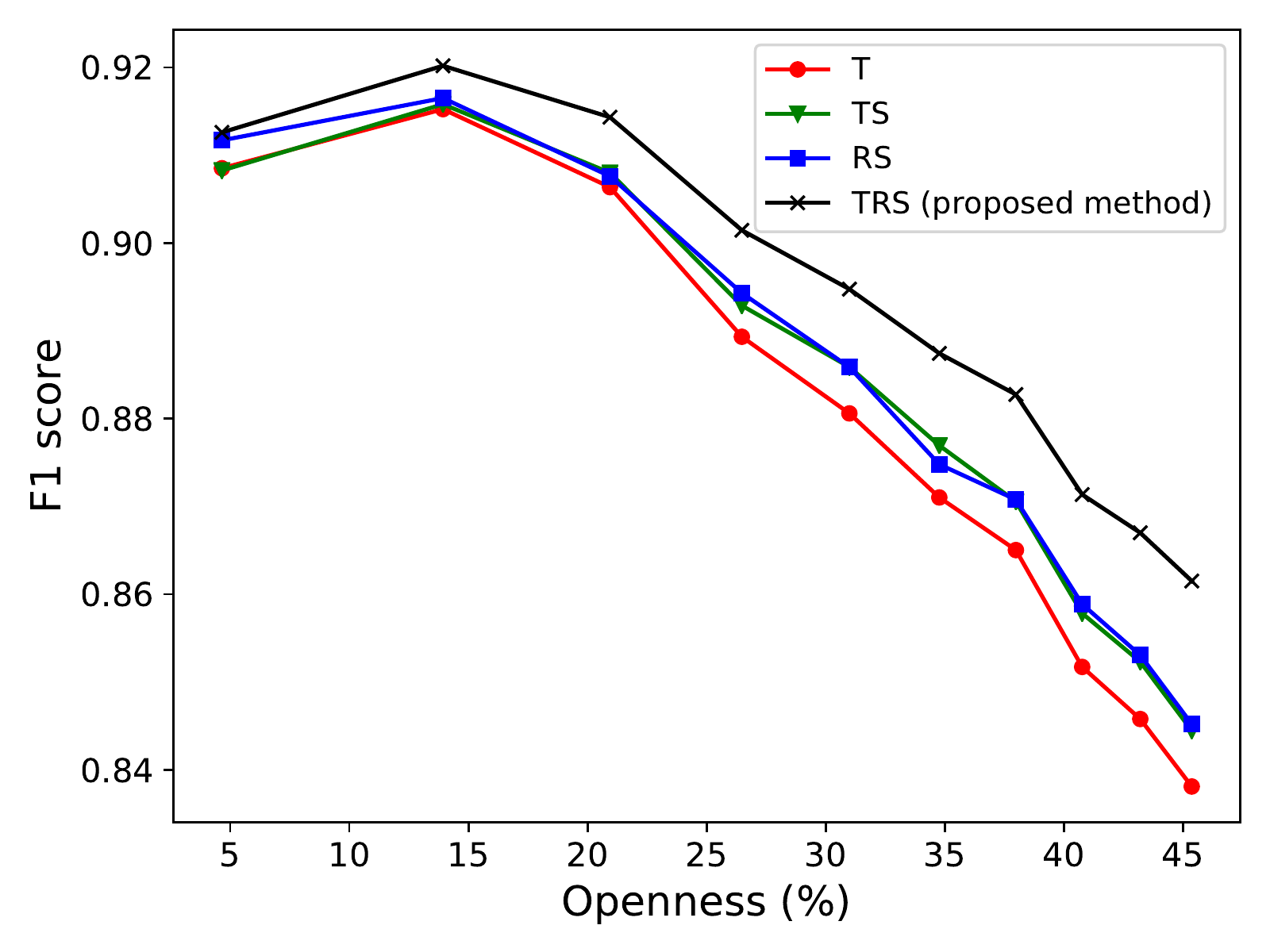}
   \caption{F1 scores against varying levels of openness achieved by four baselines in the ablation study.}
   \label{Fig_6}
\end{figure}

\cref{Fig_6} shows the F1 score changes yielded by the four baseline methods according to openness. For each openness value, we repeated a randomized unknown class sampling process five times and averaged the results, except in the case where all unknown classes were used. First, the F1 scores of all models generally decreased as openness increased. This reveals that more unknown classes or samples during testing usually leads to OSR performance degradation. Compared with T, TRS always provided better performance, and the performance gap increased as openness increased. This result demonstrates that more sophisticated decision boundaries can be established by combining the knowledge provided by the student and teacher. When the openness value was small, there were no significant differences between TRS and the other baselines. However, as the openness increased, TRS significantly outperformed not only T but also TS and RS, showing that both the teacher and the recommender are necessary in the proposed method when various types of unknown samples are likely to emerge.

\subsection{Comparison with state-of-the-art OSR methods}
In this section, the proposed method was compared with state-of-the-art methods. The proposed method was evaluated under two different experimental setups. In the first setup, the unknown detection performance was measured in terms of the area under the receiver operating characteristic curve (AUROC) metric. In the second setup, the OSR performance was measured in terms of the macroaveraged F1 score considering both known classes and unknowns.

\begin{table*}[h]
\centering
\caption{Unknown detection performance in terms of the AUROC metric. N/R indicates that a particular figure was not reported in the original work. For CIFAR+50, the standard deviation is not reported because only one open set scenario is available for evaluation.}
\label{Table_1}
\begin{tabular}{l|ccccc}
\hline								
Method & SVHN & CIFAR-10 & CIFAR+10 & CIFAR+50 & Tiny-ImageNet \\
\hline
Openness & 13.4\% & 13.4\% & 33.3\% & 62.9\% & 57.4\% \\
\hline
SoftMax&0.886$\pm$0.014&0.677$\pm$0.038&0.816$\pm$N/R&0.805&0.577$\pm$N/R\\
OpenMax (CVPR16)~\cite{Bendale2016}&0.894$\pm$0.013&0.695$\pm$0.044&0.817$\pm$N/R&0.796&0.576$\pm$N/R\\
G-OpenMax (BMVC17)~\cite{Ge2017}&0.896$\pm$0.017&0.675$\pm$0.044&0.827$\pm$N/R&0.819&0.580$\pm$N/R\\
OSRCI (ECCV18)~\cite{Neal2018}&0.910$\pm$0.010&0.699$\pm$0.038&0.838$\pm$N/R&0.827&0.586$\pm$N/R\\
C2AE (CVPR19)~\cite{Oza2019}&0.892$\pm$0.013&0.711$\pm$0.008&0.810$\pm$0.005&0.803&0.581$\pm$0.019\\
CROSR (CVPR19)~\cite{Yoshihashi2019}&0.899$\pm$0.018&N/R&N/R&N/R&0.589$\pm$N/R\\
GDFR (CVPR20)~\cite{Perera2020}&0.935$\pm$0.018&\textbf{0.807$\pm$0.039}&0.928$\pm$0.002&0.926&0.608$\pm$0.017\\
Ours&\textbf{0.937}$\pm$0.010&0.791$\pm$0.032&\textbf{0.930$\pm$0.009}&\textbf{0.928}&\textbf{0.674$\pm$0.020}\\
\hline
\end{tabular}
\end{table*}

\textbf{Unknown detection.} We carried out an unknown detection performance comparison by following the protocol described in~\cite{Neal2018} and utilizing four image datasets: SVHN~\cite{Netzer1952}, CIFAR-10~\cite{Krizhevsky2009a}, CIFAR-100 ~\cite{Krizhevsky2009a}, and Tiny-ImageNet~\cite{Le2016}. Here, Tiny-ImageNet is a subdataset of 200 classes drawn from the ImageNet dataset~\cite{Russakovsky2015} and downsampled to $32 \times 32$. In~\cite{Neal2018}, the authors generated open set scenarios by randomly selecting a fixed number of classes as knowns, while the remaining classes were considered “unknown”. This random class selection process was repeated five times, and the average AUROC was reported. To maintain the consistency of the comparison, we used the same class splits as those employed in~\cite{Neal2018}, and the results obtained based on the class splits utilized in~\cite{Perera2020} are reported for comparison purposes. Specifically, for SVHN and CIFAR-10, six classes were selected as knowns, and the remaining four classes were considered unknown. Open set scenarios defined as CIFAR+10 and CIFAR+50 were also used. In CIFAR+10, four nonanimal categories from CIFAR-10 were used as knowns, and 10 animal categories from CIFAR-100 were added as unknowns during testing. Similarly, four nonanimal categories from CIFAR-10 and 50 animal categories from CIFAR-100 were selected for CIFAR+50. Finally, 20 classes were chosen as knowns, and the remaining 180 classes were used as unknowns for Tiny-ImageNet.

\cref{Table_1} shows the results of the unknown detection performance comparison. The proposed method was marginally better on SVHN and marginally worse on CIFAR-10 than GDFR, which exhibited the best results among the comparison methods. However, for more difficult open set scenarios with larger openness levels, the proposed method outperformed the other approaches. In particular, the proposed method was able to achieve a significant improvement on the Tiny-ImageNet dataset, with a 6.6\% increase in performance.

\textbf{Open Set Recognition.} An OSR system must classify known samples effectively while also effectively rejecting unknown samples. Thus, in the experiments described below, the proposed method was compared with state-of-the-art methods by measuring their classification performance on known classes and “unknown” using macroaveraged F1 scores. The models were trained with all training samples of one dataset, but during testing, another dataset was added to the test set as “unknown”. Specifically, MNIST and CIFAR-10 were used for training.

Omniglot~\cite{Lake2015}, MNIST-Noise, and Noise were added as “unknown” when MNIST was used for training. Here, Noise involved a set of synthesized images obtained by sampling each pixel value independently from a uniform distribution on [0, 1]. MNIST-Noise is also a synthesized image dataset made by superimposing the test samples of MNIST on Noise. Each unknown dataset contained 10,000 samples, making the unknown ratio 1:1 in the test dataset. \cref{Table_2} shows the F1 scores obtained for each unknown dataset. The results show that the proposed method achieved overwhelming performance on all given unknown datasets.

\begin{table}[t]
\small
\centering
\caption{OSR results obtained on the MNIST dataset with various “unknown” datasets added to the test set. Performance was measured using macroaveraged F1 scores.}
\label{Table_2}
\begin{tabular}{l|ccc}
\hline								
\multirow{2}{*}{Method}&\multirow{2}{*}{Omniglot}&MNIST&\multirow{2}{*}{Noise}\\
 & &-Noise& \\
\hline	
SoftMax&0.592&0.641&0.826\\
OpenMax (CVPR16)~\cite{Bendale2016}&0.680&0.720&0.890\\
LadderNet+SoftMax~\cite{Yoshihashi2019}&0.588&0.772&0.828\\
LadderNet+OpenMax~\cite{Yoshihashi2019}&0.764&0.821&0.826\\
DHRNet+SoftMax~\cite{Yoshihashi2019}&0.595&0.801&0.829\\
DHRNet+OpenMax~\cite{Yoshihashi2019}&0.780&0.816&0.826\\
CROSR (CVPR19)~\cite{Yoshihashi2019}&0.793&0.827&0.826\\
CGDL (CVPR20)~\cite{Sun2020}&0.850&0.887&0.859\\
Ours&\textbf{0.907}&\textbf{0.948}&\textbf{0.948}\\
\hline
\end{tabular}
\end{table}

When CIFAR-10 was used for training, cropped or resized samples from ImageNet and the Large-Scale Scene Understanding (LSUN) dataset were used as “unknown” following the protocol suggested in~\cite{Yoshihashi2019}. Specifically, ImageNet-crop, ImageNet-resize, LSUN-crop, and LSUN-resize were added to the test set of CIFAR-10 in the testing phase. Each unknown dataset contained 10,000 samples for setting a known-to-unknown ratio of 1:1. More details on the unknown datasets can be found in~\cite{Liang2018}. \cref{Table_3} shows the comparison results obtained on CIFAR-10. In the table, the proposed method provided the second-highest performance on ImageNet-crop and ImageNet-resize, achieving slightly worse performance relative to CGDL. However, the proposed method was able to achieve massive performance improvements on LSUN-crop and LSUN-resize, revealing its overall superiority over the other methods. Considering both the MNIST and CIFAR-10 cases, we can say that the proposed method achieved a new state-of-the-art OSR performance.

\begin{table}[t]
\small
\centering
\caption{OSR results obtained on the CIFAR-10 dataset with various “unknown” datasets added to the test set. Performance was measured using macroaveraged F1 scores.}
\label{Table_3}
\begin{tabular}{p{11mm}|>{\centering\arraybackslash}p{11mm}>{\centering\arraybackslash}p{11mm}cc|c}
\hline	
\multirow{2}{*}{Method}&ImageNet&ImageNet&LSUN&LSUN&\multirow{2}{*}{Avg.}\\
 &-crop&-resize&-crop&-resize\\
 \hline
SoftMax&0.639&0.653&0.642&0.647&0.645\\
OpenMax&0.660&0.684&0.657&0.668&0.667\\
CROSR&0.721&0.735&0.720&0.749&0.731\\
CGDL&\textbf{0.840}&\textbf{0.832}&0.806&0.812&0.823\\
GDFR&0.821&0.792&\textbf{0.843}&0.805&0.813\\
Ours&0.839&0.817&\textbf{0.843}&\textbf{0.847}&\textbf{0.837}\\
\hline
\end{tabular}
\end{table}

\section{Conclusion}
\label{sec:conclusion}

In this paper, we proposed a novel OSR method by harnessing three networks: a teacher network, a student network, and a recommender network. In the proposed method, a knowledge distillation technique and a synthetic unknown class learning method were introduced. Knowledge distillation helped to make room for learning unknowns by softening the posterior probabilities of a softmax CNN. Furthermore, the synthetic unknown class learning method helped the student directly learn diverse unknown-like samples after generating them. Experiments on several benchmark datasets showed that both techniques helped improve the OSR performance of our approach. As a result, the proposed method significantly outperformed other state-of-the-art OSR methods in terms of both unknown detection performance and OSR performance. However, whether the proposed method is effective for large-scale images should be further evaluated in the next study.

\section{Supplementary Material}
\label{sec:Supp}
\subsection{Recommender network architecture}

 We applied the network architecture shown in~\cref{Table_1} for the recommender network. Here, FC($x$) is a fully connected layer with $x$ nodes. R is a reshaping layer. C($x,y,z$) and TC($x,y,z$) are a convolutional layer and a transposed convolutional layer, respectively, with $x$ filters, $y\times y$ kernels, and a stride of $z$, respectively. In the input layer, Z(100)×FC(100)($\bm{cv}$) denotes that the condition vector $\bm{cv}$ was embedded in 100 dimensions and multiplied with a 100-dimensional noise vector. Sigmoid activation was used for the output layer, and a leaky rectified linear unit (LReLU) was employed for the other layers.
 
 \newcolumntype{C}[1]{>{\centering\arraybackslash}p{#1}}
\begin{table}[h]
\centering
\caption{Recommender network architectures used for MNIST and the other datasets.}
\label{Table_1}
\begin{tabular}{C{0.5\linewidth}|C{0.37\linewidth}}
\hline\hline
\multicolumn{2}{c}{MNIST}\\
\hline									
Generator & Discriminator \\
\hline	
Input: Z(100)×FC(100)($\bm{cv}$) & Input: (28, 28, 1)\\
FC($7\times7\times128$) & C(64, 3, 2)\\
R(7, 7, 128) & C(64, 3, 2)\\
TC(128, 4, 2) & FC(1)\\
TC(128, 4, 2) & Output: 1\\
C(1, 7, 1) & \\
Output: (28, 28, 1) & \\ 
\hline\hline
\multicolumn{2}{c}{}\\
\hline\hline
\multicolumn{2}{c}{Others}\\
\hline									
Generator & Discriminator \\
\hline	
Input: Z(100)×FC(100)($\bm{cv}$) & Input: (32, 32, 3)\\
FC($4\times4\times256$) & C(64, 3, 2)\\
R(4, 4, 256) & C(128, 3, 2)\\
TC(128, 4, 2) & C(128, 3, 2)\\
TC(128, 4, 2) & C(256, 3, 2)\\
TC(128, 4, 2) & FC(1)\\
C(3, 3, 1) & Output: 1\\
Output: (32, 32, 3) & \\ 
\hline\hline
\end{tabular}
\end{table}

\subsection{Sensitivity Analysis}
To implement the proposed method, we must set two hyperparameters: the temperature $\tau$ for relaxing the posterior probabilities of the teacher network during knowledge distillation and the balancing parameter $\alpha$ for adjusting the effect of synthetic unknown samples on synthetic unknown class learning. For sensitivity analysis, the macroaveraged F1 score was measured for the case used in the ablation study, in which MNIST was used as “known” and EMNIST was added as “unknown” during testing. \cref{Fig_1} shows the sensitivities of the two hyperparameters at different openness values. The figures show that the combinations of $\alpha$ between 0.5 and 1.5 with $\tau$ between 2.5 and 5 provided high overall performance. However, with $\alpha$=0 (no unknown synthetic class learning) and $\alpha>1.5$ (when the student learns too much from the synthetic unknown samples), the performance decreased considerably, especially when a high temperature was given. In addition, excessive softening of the posterior probabilities with a high temperature can significantly degrade the performance regardless of the selected value of $\alpha$.

\begin{figure}[h]
  \centering
  \begin{subfigure}{0.7\linewidth}
    \includegraphics[width=\linewidth]{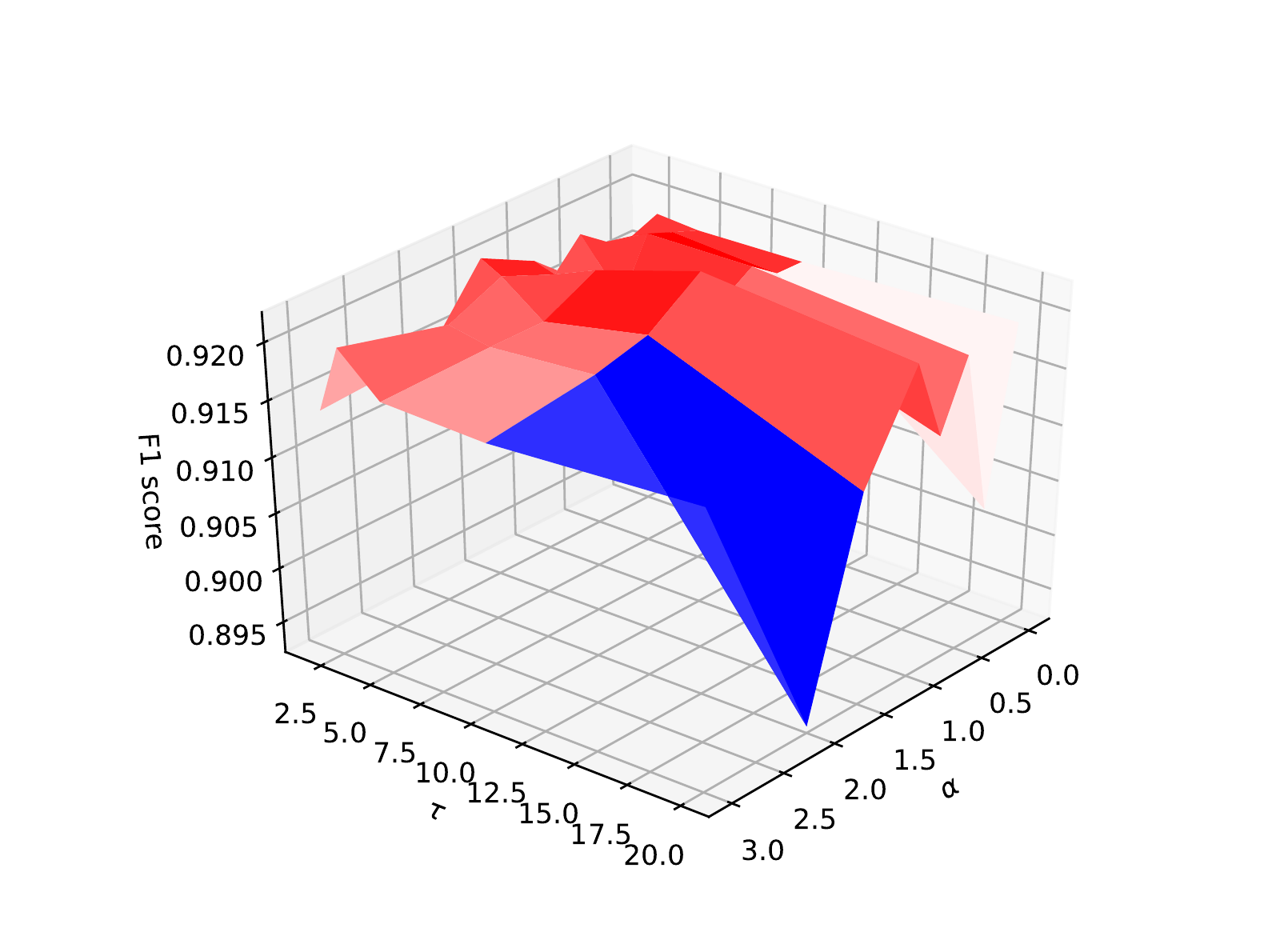}
    \caption{Openness: 13.93\%}
  \end{subfigure}
  \hfill
  \begin{subfigure}{0.7\linewidth}
    \includegraphics[width=\linewidth]{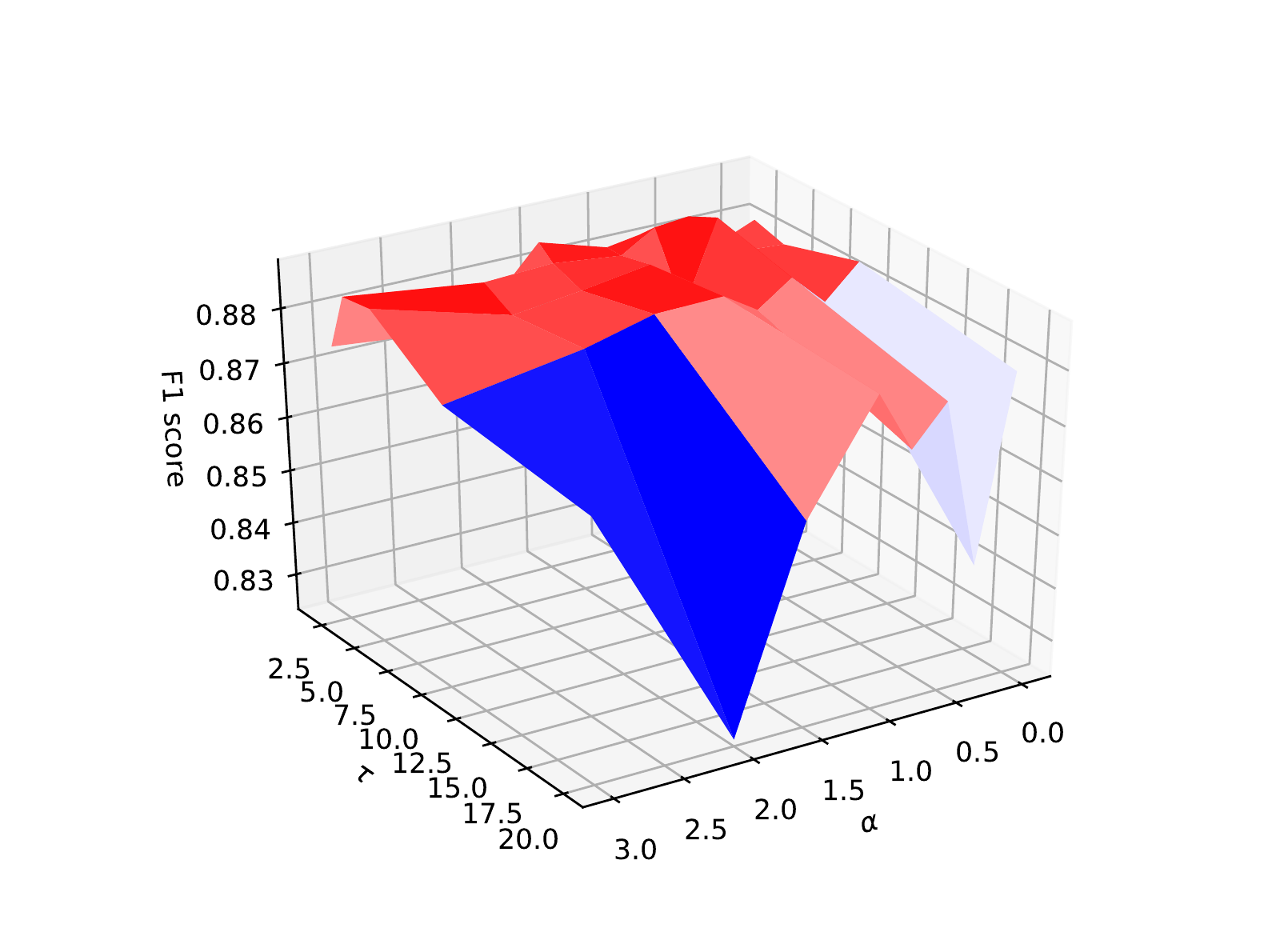}
    \caption{Openness: 34.77\%}
  \end{subfigure}
    \begin{subfigure}{0.7\linewidth}
    \includegraphics[width=\linewidth]{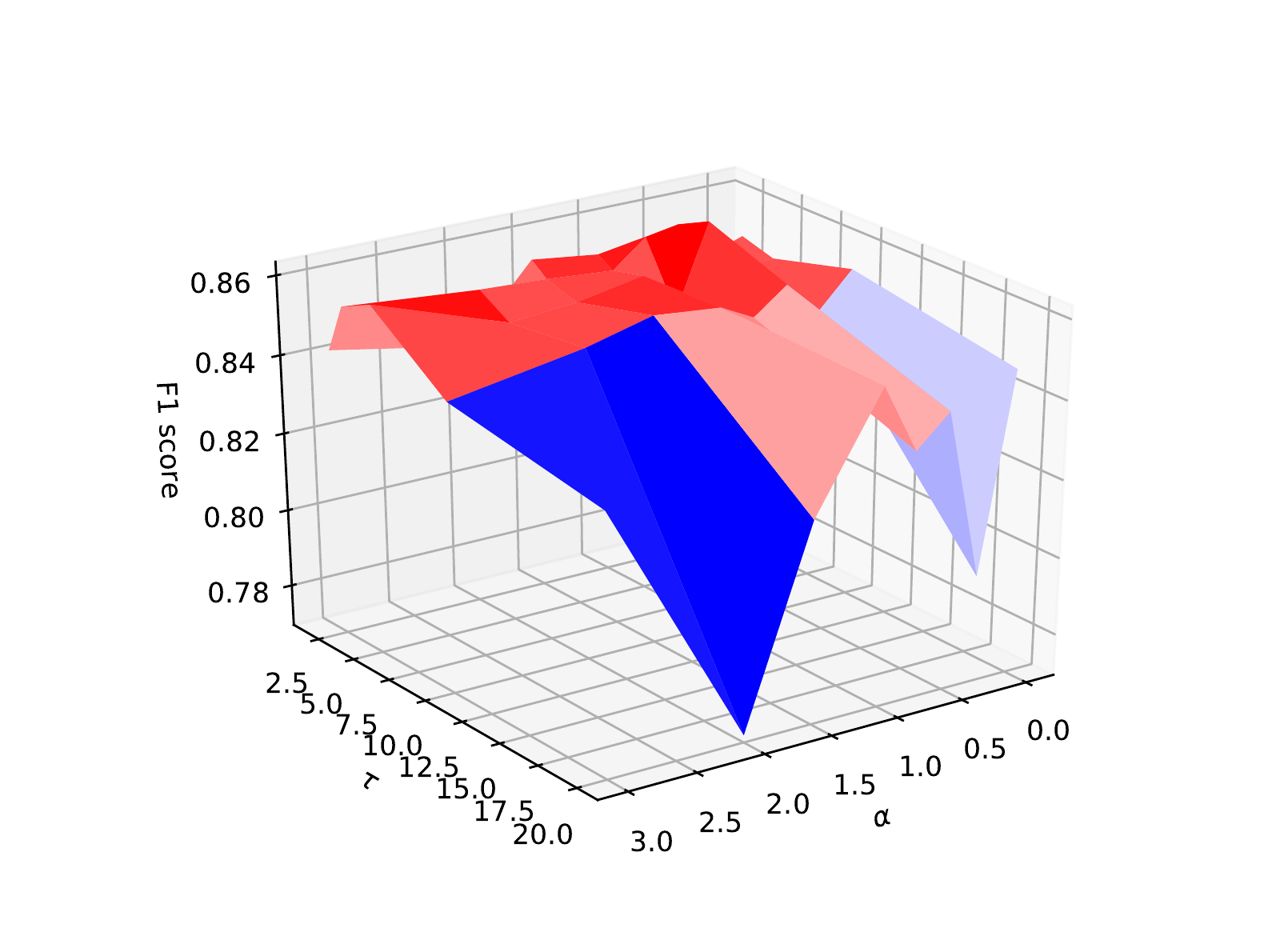}
    \caption{Openness: 45.36\%}
  \end{subfigure}
  \caption{Sensitivity of the proposed method to different hyperparameters with varying openness levels.}
  \label{Fig_1}
\end{figure}

{\small
\bibliographystyle{ieee_fullname}
\bibliography{egbib}

\begin{thebibliography}{10}\itemsep=-1pt

\bibitem{Bendale2016}
Abhijit Bendale and Terrance~E. Boult.
\newblock {Towards open set deep networks}.
\newblock In {\em IEEE/CVF conference on computer vision and pattern
  recognition (CVPR)}, pages 1563--1572, 2016.

\bibitem{Cohen2017}
Gregory Cohen, Afshar Saeed, Jonathan Tapson, and Andre van Schaik.
\newblock {EMNIST: Extending MNIST to handwritten letters}.
\newblock In {\em International Joint Conference on Neural Networks (IJCNN)},
  pages 2921--2926, 2017.

\bibitem{Ge2017}
Zongyuan Ge, Sergey Demyanov, and Rahil Garnavi.
\newblock {Generative OpenMax for multi-class open set classification}.
\newblock In {\em British Machine Vision Association (BMVC)}, 2017.

\bibitem{He2016}
Kaiming He, Xiangyu Zhang, Shaoqing Ren, and Jian Sun.
\newblock {Deep residual learning for image recognition}.
\newblock In {\em IEEE/CVF conference on computer vision and pattern
  recognition (CVPR)}, pages 770--778, 2016.

\bibitem{Hendrycks2017}
Dan Hendrycks and Kevin Gimpel.
\newblock {A baseline for detecting misclassified and out-of-distribution
  examples in neural networks}.
\newblock In {\em International Conference on Learning Representations (ICLR)},
  pages 1--12, 2017.

\bibitem{Henrydoss2017}
James Henrydoss, Steve Cruz, Ethan~M. Rudd, Manuel Gunther, and Terrance~E.
  Boult.
\newblock {Incremental open set intrusion recognition using extreme value
  machine}.
\newblock In {\em IEEE International Conference on Machine Learning and
  Applications (ICMLA)}, pages 1089--1093, 2017.

\bibitem{Hinton2015}
Geoffrey Hinton, Oriol Vinyals, and Jeff Dean.
\newblock {Distilling the knowledge in a neural network}.
\newblock In {\em Advances in Neural Information Processing Systems (NIPS)-Deep
  Learning Workshop}, 2015.

\bibitem{Jang2020b}
Jaeyeon Jang and Chang~Ouk Kim.
\newblock {Collective decision of one-vs-rest networks for open set
  recognition}.
\newblock {\em arXiv:2103.10230}, 2021.

\bibitem{Jang2021}
Jaeyeon Jang and Chang~Ouk Kim.
\newblock {Teacher-explorer-student learning: A novel learning method for open
  set recognition}.
\newblock {\em arXiv preprint arXiv: 2103.12871}, 2021.

\bibitem{Jang2020a}
Jaeyeon Jang, Minkyung Seo, and Chang~Ouk Kim.
\newblock {Support weighted ensemble model for open set recognition of wafer
  map defects}.
\newblock {\em IEEE Transactions on Semiconductor Manufacturing},
  33(4):635--643, 2020.

\bibitem{Kardan2017}
Navid Kardan and Kenneth~O. Stanley.
\newblock {Mitigating fooling with competitive overcomplete output layer neural
  networks}.
\newblock In {\em International Joint Conference on Neural Networks (IJCNN)},
  pages 518--525, 2017.

\bibitem{Kawaguchi2017}
Kenji Kawaguchi, Leslie~Pack Kaelbling, and Yoshua Bengio.
\newblock {Generalization in deep learning}.
\newblock {\em arXiv:1710.05468}, 2017.

\bibitem{Krizhevsky2009a}
Alex Krizhevsky, Vinod Nair, and Geoffrey Hinton.
\newblock {Cifar-10 and cifar-100 datasets.}, 2009.

\bibitem{Lake2015}
Brenden~M. Lake, Ruslan Salakhutdinov, and J.~B. Tenenbaum.
\newblock {Human-level concept learning through probabilistic program
  induction}.
\newblock {\em Science}, 350(6266):1332--1338, 2015.

\bibitem{Le2016}
Ya Le and Xuan Yang.
\newblock {Tiny ImageNet visual recognition challenge}.
\newblock {\em CS 231N}, 2015.

\bibitem{Web2012}
{Li Deng}.
\newblock {The MNIST database of handwritten digit images for machine learning
  research}.
\newblock {\em IEEE Signal Processing Magazine}, 29(6):141--142, 2012.

\bibitem{Liang2018}
Shiyu Liang, Yixuan Li, and Rayadurgam Srikant.
\newblock {Enhancing the reliability of out-of-distribution image detection in
  neural networks}.
\newblock In {\em International Conference on Machine Learning (ICML)}, pages
  1--15, 2018.

\bibitem{Moosavi-Dezfooli2019}
Seyed-Mohsen Moosavi-Dezfooli, Alhussein Fawzi, Jonathan Uesato, and Pascal
  Frossard.
\newblock {Robustness via curvature regularization, and vice versa}.
\newblock In {\em IEEE/CVF conference on computer vision and pattern
  recognition (CVPR)}, pages 9070--9078. IEEE, 2019.

\bibitem{Neal2018}
Lawrence Neal, Matthew Olson, Xiaoli Fern, Weng-Keen Wong, and Fuxin Li.
\newblock {Open set learning with counterfactual images}.
\newblock In {\em European Conference on Computer Vision (ECCV)}, pages
  613--628, 2018.

\bibitem{Netzer1952}
Yuval Netzer, Tao Wang, Adam Coates, Alessandro Bissacco, Bo Wu, and Andrew~Y.
  Ng.
\newblock {Reading digits in natural images with unsupervised feature
  learning}.
\newblock In {\em Advances in Neural Information Processing Systems
  (NIPS)-Workshop on Deep Learning and Unsupervised Feature Learning}, 2011.

\bibitem{Oza2019}
Poojan Oza and Vishal~M Patel.
\newblock {C2AE: Class conditioned auto-encoder for open-set recognition}.
\newblock In {\em IEEE/CVF conference on computer vision and pattern
  recognition (CVPR)}, pages 2307--2316, 2019.

\bibitem{Perera2020}
Pramuditha Perera, Vlad~I. Morariu, Rajiv Jain, Varun Manjunatha, Curtis
  Wigington, Vicente Ordonez, and Vishal~M. Patel.
\newblock {Generative-Discriminative Feature Representations for Open-Set
  Recognition}.
\newblock In {\em IEEE/CVF Conference on Computer Vision and Pattern
  Recognition (CVPR)}, pages 11811--11820, 2020.

\bibitem{Rocha2017}
Anderson Rocha, Walter~J. Scheirer, Christopher~W. Forstall, Thiago Cavalcante,
  Antonio Theophilo, Bingyu Shen, Ariadne R.~B. Carvalho, and Efstathios
  Stamatatos.
\newblock {Authorship attribution for social media forensics}.
\newblock {\em IEEE Transactions on Information Forensics and Security},
  12(1):5--33, 2017.

\bibitem{Russakovsky2015}
Olga Russakovsky, Jia Deng, Hao Su, Jonathan Krause, Sanjeev Satheesh, Sean Ma,
  Zhiheng Huang, Andrej Karpathy, Aditya Khosla, Michael Bernstein,
  Alexander~C. Berg, and Li Fei-Fei.
\newblock {ImageNet large scale visual recognition challenge}.
\newblock {\em International Journal of Computer Vision}, 115(3):211--252,
  2015.

\bibitem{Scheirer2013}
Walter~J. Scheirer, A. {de Rezende Rocha}, Archana Sapkota, and Terrance~E.
  Boult.
\newblock {Toward open set recognition}.
\newblock {\em IEEE Transactions on Pattern Analysis and Machine Intelligence},
  35(7):1757--1772, 2013.

\bibitem{Scheirer2014}
Walter~J. Scheirer, Lalit~P. Jain, and Terrance~E. Boult.
\newblock {Probability models for open set recognition}.
\newblock {\em IEEE Transactions on Pattern Analysis and Machine Intelligence},
  36(11):2317--2324, 2014.

\bibitem{Shu2017}
Lei Shu, Hu Xu, and Bing Liu.
\newblock {DOC: Deep open classification of text cocuments}.
\newblock In {\em Conference on Empirical Methods in Natural Language
  Processing (EMNLP)}, pages 2911--2916, 2017.

\bibitem{Simonyan2015}
Karen Simonyan and Andrew Zisserman.
\newblock {Very deep convolutional networks for large-scale image recognition}.
\newblock In {\em International Conference on Learning Representations (ICLR)},
  2015.

\bibitem{Spigler2019}
Giacomo Spigler.
\newblock {Denoising autoencoders for overgeneralization in neural networks}.
\newblock {\em IEEE Transactions on Pattern Analysis and Machine Intelligence},
  42(4):998 -- 1004, 2019.

\bibitem{Sun2020}
Xin Sun, Zhenning Yang, Chi Zhang, Keck-Voon Ling, and Guohao Peng.
\newblock {Conditional gaussian distribution learning for open set
  recognition}.
\newblock In {\em IEEE/CVF conference on computer vision and pattern
  recognition (CVPR)}, pages 13477--13486, 2020.

\bibitem{Tan2019}
Mingxing Tan and Quoc~V. Le.
\newblock {EfficientNet: Rethinking model scaling for convolutional neural
  networks}.
\newblock In {\em International Conference on Machine Learning (ICML)}, pages
  10691--10700, 2019.

\bibitem{Wong2019}
Kelvin Wong, Shenlong Wang, Mengye Ren, Ming Liang, and Raquel Urtasun.
\newblock {Identifying unknown instances for autonomous driving}.
\newblock In {\em Conference on Robot Learning (CoRL)}, pages 384--393, 2020.

\bibitem{Yang2020}
Hong-Ming Yang, Xu-Yao Zhang, Fei Yin, Qing Yang, and Cheng-Lin Liu.
\newblock {Convolutional prototype network for open set recognition}.
\newblock {\em IEEE Transactions on Pattern Analysis and Machine Intelligence},
  pages 1--13, 2020.

\bibitem{Yoshihashi2019}
Ryota Yoshihashi, Wen Shao, Rei Kawakami, Shaodi You, Makoto Iida, and Takeshi
  Naemura.
\newblock {Classification-reconstruction learning for open-set recognition}.
\newblock In {\em IEEE/CVF conference on computer vision and pattern
  recognition (CVPR)}, pages 4011--4020, 2019.

\end{thebibliography}
}

\end{document}